\documentclass[conference]{IEEEtran}
\renewcommand\IEEEkeywordsname{Keywords}
\usepackage{graphicx}
\usepackage{amsmath}
\usepackage{array}
\usepackage{multirow}
\usepackage{makecell}
\usepackage{hyperref} 
\usepackage{fancyhdr}
\usepackage[caption=false,font=footnotesize]{subfig} 
\ifCLASSINFOpdf
\else
\fi
\begin{document}
\fancypagestyle{firstpage}{
    \fancyhf{}
    \fancyfoot[L]{\footnotesize (979-8-3315-1127-2/24/\$31.00~\copyright~2024 IEEE)}
	\renewcommand{\headrulewidth}{0pt}
}

%
\title{FedBrain-Distill: Communication-Efficient Federated Brain Tumor Classification Using Ensemble Knowledge Distillation on Non-IID Data}

\author{\IEEEauthorblockN{Rasoul Jafari Gohari}
\IEEEauthorblockA{Department of Computer Science\\
Shahid Bahonar University of Kerman\\
Kerman, Iran\\
Email: rjafari@math.uk.ac.ir}
\and
\IEEEauthorblockN{Laya Aliahmadipour}
\IEEEauthorblockA{Department of Computer Science\\
Shahid Bahonar University of Kerman\\
Kerman, Iran\\
Email: l.aliahmadipour@uk.ac.ir}
\and
\IEEEauthorblockN{Ezat Valipour}
\IEEEauthorblockA{Department of Applied Mathematics\\
Shahid Bahonar University of Kerman\\
Kerman, Iran\\
Email: valipour@uk.ac.ir}}


%


\maketitle
\thispagestyle{firstpage}

\begin{abstract}
Brain is one the most complex organs in the human body. Due to its complexity, classification of brain tumors still poses a significant challenge, making brain tumors a particularly serious medical issue. Techniques such as Machine Learning (ML) coupled with Magnetic Resonance Imaging (MRI) have paved the way for doctors and medical institutions to classify different types of tumors. However, these techniques suffer from limitations that violate patients' privacy. Federated Learning (FL) has recently been introduced to solve such an issue, but the FL itself suffers from limitations like communication costs and dependencies on model architecture, forcing all models to have identical architectures. In this paper, we propose FedBrain-Distill, an approach that leverages Knowledge Distillation (KD) in an FL setting that maintains the users privacy and ensures the independence of FL clients in terms of model architecture. FedBrain-Distill uses an ensemble of teachers that distill their knowledge to a simple student model. The evaluation of FedBrain-Distill demonstrated high-accuracy results for both Independent and Identically Distributed (IID) and non-IID data with substantial low communication costs on the real-world Figshare brain tumor dataset. It is worth mentioning that we used Dirichlet distribution to partition the data into IID and non-IID data. All the implementation details are accessible through our Github repository\footnote{\url{https://github.com/russelljeffrey/FedBrain-Distill}}.
\\
\end{abstract}
\begin{IEEEkeywords}
Federated Learning, Knowledge Distillation, Brain Tumor Classification, non-IID data
\end{IEEEkeywords}


%
\IEEEpeerreviewmaketitle

\section{Introduction}
Recently, the classification of brain tumors has become a top priority due to its considerable influence on patient survival rate. This phenomenon can lead to irreparable neurological damage and in some cases mortality if the medical diagnosis is not accurate. Brain tumors are one of the most challenging medical issues whose frequency of occurrence has rapidly grown among medical patients. Numerous factors differentiate one tumor from another, some of which include tumor’s size, its growth pattern and its malignancy.
\\
In light of technological advancements in both medical imaging and machine learning (ML), techniques such as Magnetic Resonance Imaging (MRI) can be easily leveraged by ML algorithms such as Convolutional Neural Networks (CNNs). These ML models have illustrated their importance in accurate brain tumor classification. Yet, when it comes to the deployment of such models, we face unintended consequences. Firstly, patients’ privacy is violated due to sharing users' sensitive data among multiple ML models. In other words, data is distributed among as many models as there is to be trained. Secondly, we are ignoring the computational complexities of CNN models in order to obtain high accuracy. Thirdly, even if we do not take these two hurdles into consideration, some medical institutions may require a different architecture when using CNN models.
\\
Knowledge Distillation (KD) has recently emerged as an architecture-agnostic solution that stands out due to its capability in transferring knowledge from one complex teacher model to a simple student model \cite{c1}. This technique can also be leveraged within Federated Learning (FL) settings, which is called Federated Knowledge Distillation (FKD) \cite{c2}. As a result, the FKD will be able to resolve the aforementioned challenges. To begin with, FKD maintains the privacy of users’ sensitive data by building upon FL architecture. In addition, unlike vanilla FL where clients’ architecture is similar, the FKD technique enables clients to be independent from one another when it comes to model architecture. Consequently, knowledge is distilled from a single or an ensemble of complex teachers to a simpler student model, ensuring the patients’ sensitive data is safeguarded and  computational load is handled by complex teachers that have the capacity and resources to deal with a large amount of data. Also, since one of the main criteria in FL settings is the communication cost, FKD ensures that communication occurs as optimally as possible among clients given that both teachers and student models only communicate soft labels, rather than model parameters \cite{c3}.
\\
In this paper, we propose a novel communication-efficient approach for brain tumor classification task dubbed as \textbf{FedBrain-Distill}, which distills knowledge from an ensemble of complex teachers to a simpler student model. We take advantage of multiple pre-trained VGGNet16 models for training the teachers. In addition, we take advantage of Dirichlet distribution to create non-Independent and Identically Distributed (non-IID) data among the teachers. For the FedBrain-Distill, we utilize two different scenarios in which the knowledge of 2 and 5 teachers is distilled into a simple student model. These two settings are tested in our experimentation phase under two different IID and non-IID data distributions among the teachers. This allows us to see the efficiency of FedBrain-Distill when data is highly skewed towards specific classes of data. For our experimentation, we leverage Figshare Brain Tumor dataset, that consists of 3 classes, namely meningioma, glioma and pituitary tumor. 
\\
The structure of this paper is as follows: in section II we discuss the related works that put forward similar methods. In section III, we go over the architecture of FedBrain-Distill and discuss how knowledge is distilled from teachers and finally in section IV experimental results are provided.


\section{Related Work}
Numerous studies have worked towards brain tumor classification using pretrained models. Although not based on KD technique, work of Deepak and Ameer  \cite{c4} leveraged the GoogLeNet model to extract features from brain MRI images. In fact, their work integrates multiple deep CNN models along with GoogLeNet to achieve high accuracy on Figshare MRI dataset. Wu et al. \cite{c5} proposed FedKD, which addresses the significant communication overheads inherent in traditional FL methods. They proposed leveraging adaptive mutual knowledge distillation combined with dynamic gradient compression to reduce communication overhead without sacrificing the model performance. Rahimpour et al. \cite{c6} proposed a cross-modal distillation approach that involves training a teacher model with multi-sequence MRI data and a student model with single-sequence data. They used the BraTS 2018 dataset and an in-house dataset using a customized U-net model. ShakibKhan et al. in \cite{c7} and Adepu et al. in \cite{c8} leveraged DL models to deploy teachers in order to detect and classify melanoma, respectively. Qin et al. \cite{c9} proposed a novel architecture that systematically constructs a holistic structure for transferring segmentation knowledge of medical images from teacher to student. Wang et al. \cite{c10} leveraged conditional probability representations to extract knowledge from the teacher model to do multi-task segmentation of abdominal organs. Wang et al. \cite{c11} proposed a novel method that reduces the impact of inaccurate predictions from less relevant local models, thereby enhancing the overall performance of the aggregated model. Work of Zhang et al. \cite{c12} addresses the challenge of data heterogeneity by using a generator model for the teachers in order to create synthetic data and distill the models’ knowledge and send it to the aggregator. Chen et al. \cite{c13} proposed MetaFed, a framework created for FL among different federations without a central server, utilizing cyclic knowledge distillation to accumulate common knowledge and achieve personalized models for each federation. And finally, Viet et al. \cite{c14} took advantage of multiple pretrained models to classify brain tumors using Figshare dataset in an FL setting. They achieved remarkable accuracy using VGGNet16 on both IID and non-IID data. 

\section{Proposed Method}
The emergence of the FL paradigm has opened doors to the research community so that instead of bringing the data to the code, the code is brought to the data from a single server. In other words, a central server called the aggregator coordinates the learning of a single global model whose parameters are gathered from a federation of clients \cite{c15}. Figure 1 shows this procedure.
\begin{figure}[htbp]
  \centering
  \includegraphics[width=\linewidth]{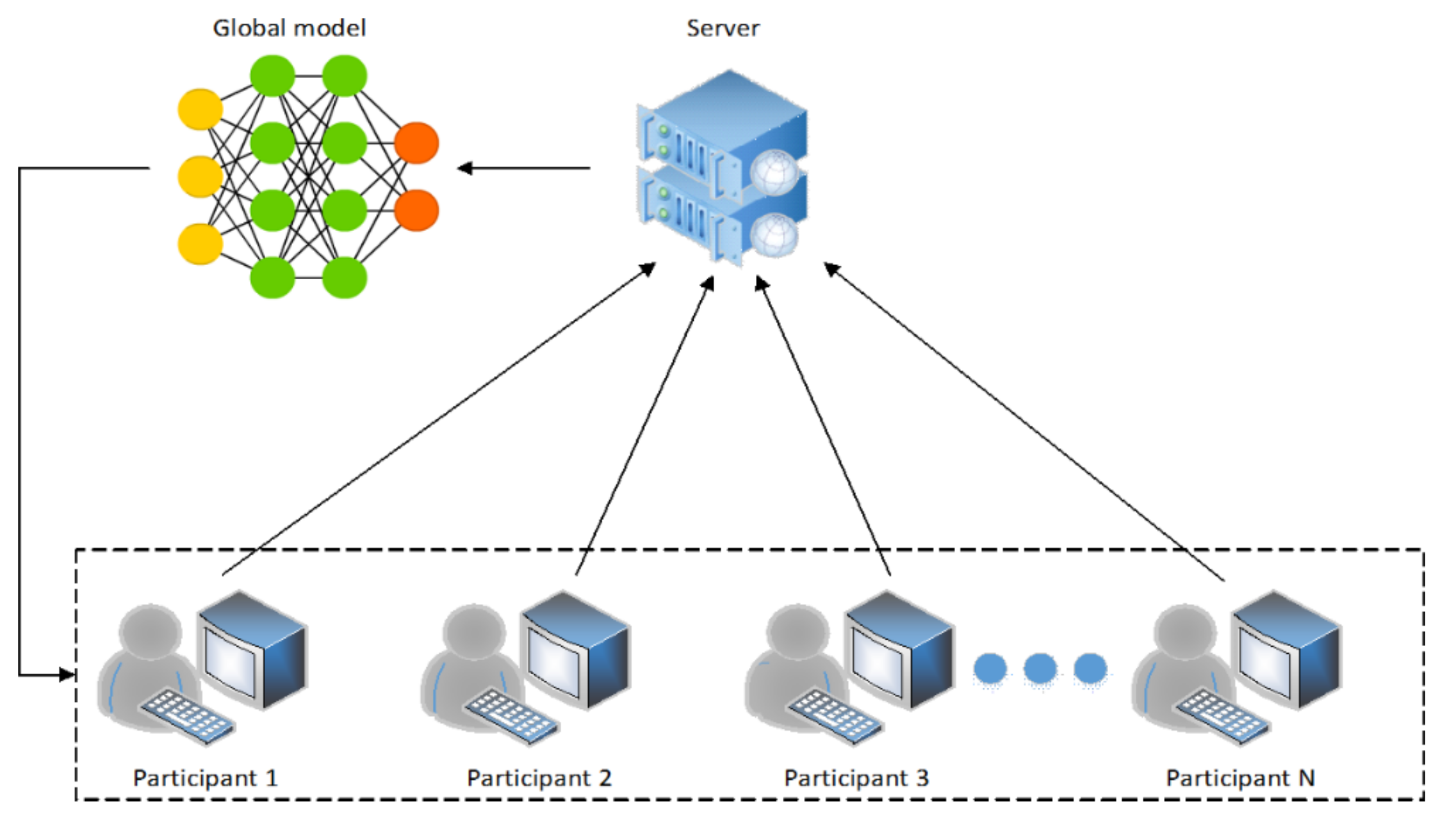}
  \caption{Federated learning overview, showing the aggregation of clients' weights with the global model in the aggregator/server.\cite{c15}.}
  \label{fig:example}
\end{figure}
FedBrain-Distill is built upon the architecture of FL. FedBrain-Distill distills the knowledge of an ensemble of teachers to a simple student model. This simple model can later be deployed on any client that may not possess the necessary hardware for training. This point of view is equivalent to the same aggregator-client relationship in FL settings. Therefore, teachers in FedBrain-Distill are the clients while the student model resides at the top in the aggregator. Figure 2 illustrates the overall architecture of FedBrain-Distill. Below, we will discuss each part in details.
\subsection{Preprocessing}
The success of both teachers and student models hinges on data preprocessing, as it unlocks rich brain image features for effective training. For this reason, all tumor images were normalized, reshaped and finally enhanced using Contrast-Limited Adaptive Histogram Equalization (CLAHE) technique \cite{c15}. Consequently, the tumor region (tumor mask) in each image is enhanced, allowing the models to extract specific tumor regions much more efficiently. Figure 3 demonstrates the difference between a normalized image without enhancement and an enhanced image before reshaping. All tumor images were of shape 512 × 512. FedBrain-Distill reshapes all images into 224 × 224 × 3. The reason behind this was that the teacher model's input layer only accepts the latter shape since the VGGNet16 is the chosen model to be used on all teachers.

\begin{figure*}[t]
  \centering
  \includegraphics[width=\textwidth]{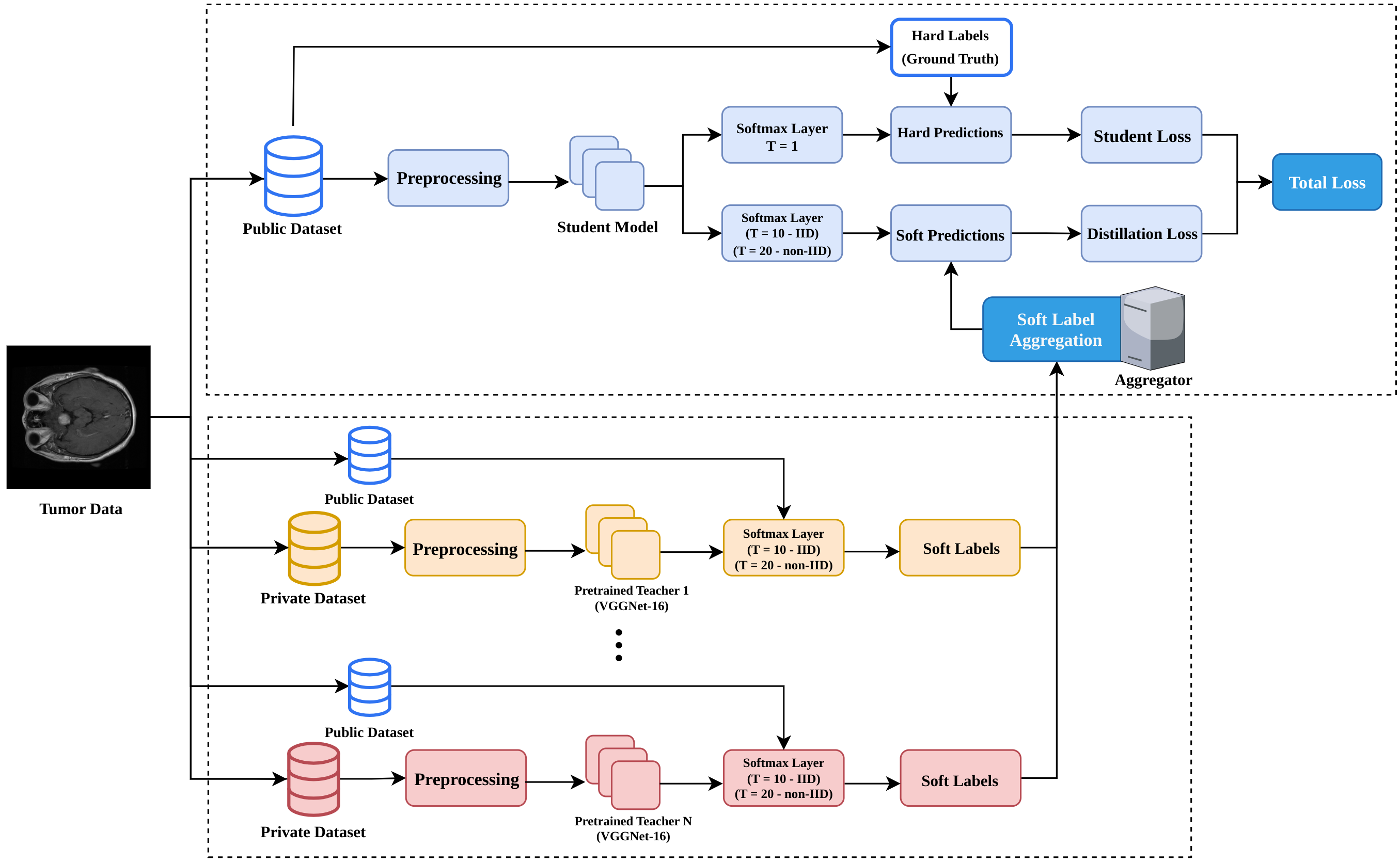}
  \caption{FedBrain-Distill workflow: The student model uses public dataset and resides in the aggregator while the teacher models use their private dataset to share their knowledge with the student using knowledge distillation. The aggregated soft labels obtained from teachers' softmax layers are distilled into the student model using Kullback-Leiber divergence. The more rounds of training there are, the less the divergence between the student and the teachers becomes.}
  \label{fig:example}
\end{figure*}

\begin{figure}[htbp]
  \centering
  \includegraphics[width=\linewidth]{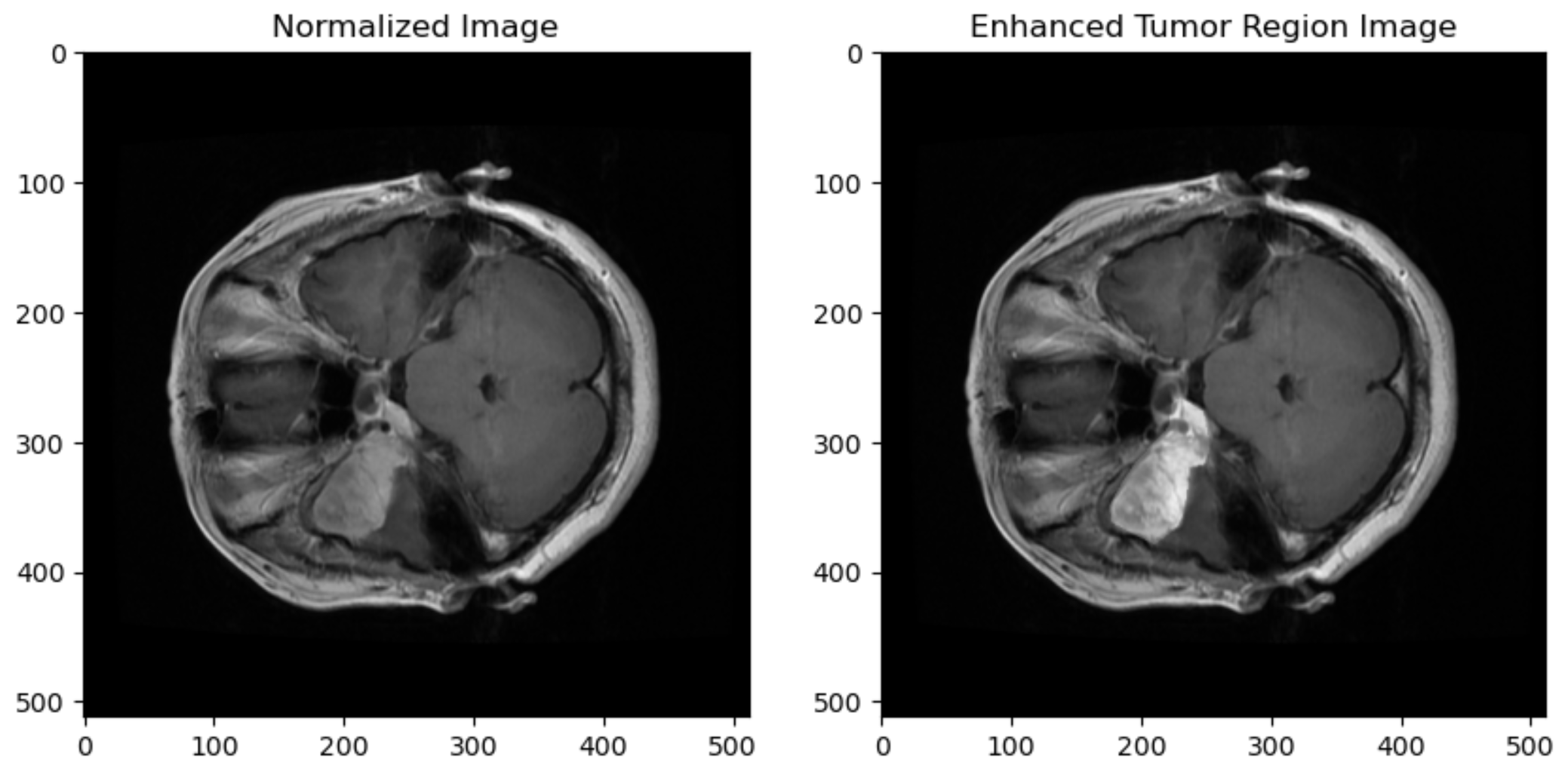}
  \caption{Comparison of normalized and enhanced tumor image before and after applying CLAHE technique.}
  \label{fig:example}
\end{figure}
\raggedbottom
\subsection{Data Partitioning}
FedBrain-Distill evaluation was carried out on both IID and non-IID data to ensure about the student results as well as its reliability on heterogeneous data. IID and non-IID data distributions were achieved in FedBrain-Distill by sampling class priors from a Dirichlet distribution, where a parameter $\alpha$ determines the heterogeneity of splits \cite{c17}. As $\alpha$ approaches 0, the partitions become more heterogeneous, whereas as $\alpha$ approaches infinity, the partitions become more uniform. FedBrain-Distill uses $\alpha = 10000$ to represent the IID setting while for the non-IID setting, FedBrain-Distill uses $\alpha = 0.5$. 

\subsection{Public and Private Dataset}
Since privacy is of utmost importance within FL settings, all clients tend to rely on their own local dataset. This dataset within the FKD setting is called a private dataset. On the other hand, the ultimate purpose of FKD is to distill knowledge from teachers to a student based on a shared dataset, which demands a second dataset that can be utilized among all participants in the entire federation. This dataset is called the public dataset that is shared among both the teachers and the student model during the distillation phase \cite{c3}. 
\subsection{Teacher Models}
Each teacher model \(\theta_t\) is trained on its own private data \(D_t\) with corresponding labels \(L_t\). The final objective for each teacher model is to minimize its local cross-entropy loss function:
\[
\mathcal{L}_t(\theta_t) = \frac{1}{|D_t|} \sum_{(x_i, y_i) \in (D_t, L_t)} \text{CrossEntropy}(f_{\theta_t}(x_i), y_i),
\]
where \(f_{\theta_t}(x_i)\) represents the output logits of the teacher model \(\theta_t\) for input \(x_i\). The optimal teacher model parameters are:
\[
\theta_t^* = \arg \min_{\theta_t} \mathcal{L}_t(\theta_t).
\]
Once each teacher model \(\theta_t^*\) is trained, soft labels are generated on a common public dataset \(X_{\text{public}}\):
\[
P_t = \sigma_{T}(f_{\theta_t^*}(X_{\text{public}})),
\]
where \(\sigma_T\) denotes the softmax function with temperature \(T\) \cite{c18}:
\[
q_i = \frac{e^{z_k / T}}{\sum_{j} e^{z_j / T}}.
\]
This softmax function is used in Neural Networks (NN) whose outcome is generally class probabilities via leveraging a softmax layer. The responsibility of this layer is to convert logits $z_k$ into probabilities $q_i$, which we call soft labels.  The temperature \(T\) controls the softness of the probability distribution. The higher the \(T\), the softer the distribution. As far as the student's model on hard labels is concerned, the parameter \(T\) is set to 1. Plus, the parameter \(T\) is set to 10 and 20 for IID and non-IID data partitioning schemes, respectively. The reasoning behind this is that when dealing with non-IID data, using a higher temperature can smooth out the noisy soft labels from different teachers. The softer distributions help the student model learn a more generalized representation of the knowledge from an ensemble of teacher models. Once all the soft labels from all teacher models are generated, they are aggregated on the aggregator by averaging the soft labels from all teachers:
\[
P_{\text{agg}} = \frac{1}{T} \sum_{t=1}^{T} P_t,
\]
where \(T\) is the total number of teacher models.

\subsection{Student Model}
The student model \(\theta_s\) is trained using a combination of the distillation loss (with the aggregated soft labels) and the student loss (with true labels from the public dataset). The total loss for training the student model is:
\[
\mathcal{L}_{\text{total}}(\theta_s) = \alpha \mathcal{L}_{\text{student}}(\theta_s) + (1 - \alpha) \mathcal{L}_{\text{distill}}(\theta_s),
\]
where $\alpha$ is a hyperparameter, allowing the student model to set a ratio for both the student loss and the distillation loss. FedBrain-Distill uses $\alpha = 0.1$ so that it is influenced by 90\% of the distillation loss. We must note that a distinction must be made between this parameter and the Dirichlet distribution $\alpha$ parameter. The student loss is the cross-entropy loss with the true labels from the public dataset:
\[
\mathcal{L}_{\text{student}}(\theta_s) = \frac{1}{N_{\text{public}}} \sum_{i=1}^{N_{\text{public}}} \text{CrossEntropy}(f_{\theta_s}(x_{\text{public}}, i), y_{\text{public}}, i),
\]
where \(N_{\text{public}}\) is the number of samples in the public dataset, and \(y_{\text{public}}\) are the true labels. The distillation loss is the Kullback-Leibler (KL) divergence between the aggregated soft labels and the student model's predictions on the public dataset:
\[
\mathcal{L}_{\text{distill}}(\theta_s) = \frac{1}{N_{\text{public}}} \sum_{i=1}^{N_{\text{public}}} \text{KL}(\sigma_T(P_{\text{agg}, i}) \| \sigma_T(f_{\theta_s}(x_{\text{public}}, i))).
\]
The student model is optimized to minimize the total loss:
\[
\theta_s^* = \arg \min_{\theta_s} \mathcal{L}_{\text{total}}(\theta_s).
\]
Ultimately, by minimizing \(\mathcal{L}_{\text{total}}\), the student model effectively learns from the rich information provided by the aggregated soft labels, capturing the knowledge distilled from multiple teacher models.

\subsection{Student Model Architecture}
We used VGGNet16 for all the teachers. The VGGNet16 has 138 million parameters \cite{c19}. Table 1 shows a summary of the architecture of our student model, which as it can be seen, has only 95434 parameters. Compared to the VGGNet16, our model is far simpler.
\vspace{-5pt}
\begin{table}[h]
\renewcommand{\arraystretch}{1.3}
\caption{Summary of the "student" Model}
\label{model_summary}
\centering
\begin{tabular}{|>{\centering\arraybackslash}p{0.5cm}|>{\centering\arraybackslash}p{2.9cm}|>{\centering\arraybackslash}p{2.4cm}|>{\centering\arraybackslash}p{1.1cm}|}
\hline
\textbf{\#} & \textbf{Layer (type)} & \textbf{Output Shape} & \textbf{Param \#} \\
\hline
1 & Conv2D & (112 × 112 × 32) & 896 \\
\hline
2 & BatchNormalization & (112 × 112 × 32) & 128 \\
\hline
3 & LeakyReLU & (112 × 112 × 32) & 0 \\
\hline
4 & MaxPooling2D & (56 × 56 × 32) & 0 \\
\hline
5 & Conv2D & (28 × 28 × 64) & 18496 \\
\hline
6 & BatchNormalization & (28 × 28 × 64) & 256 \\
\hline
7 & LeakyReLU & (28 × 28 × 64) & 0 \\
\hline
8 & MaxPooling2D & (14 × 14 × 64) & 0 \\
\hline
9 & Conv2D & (7 × 7 × 128) & 73856 \\
\hline
10 & BatchNormalization & (7 × 7 × 128) & 512 \\
\hline
11 & LeakyReLU & (7 × 7 × 128) & 0 \\
\hline
12 & MaxPooling2D & (4 × 4 × 128) & 0 \\
\hline
13 & GlobalAveragePooling2D & (128) & 0 \\
\hline
14 & Dense & (10) & 1290 \\
\hline
\end{tabular}

\vspace{0.05cm}
\begin{tabular}{|>{\centering\arraybackslash}p{3.85cm}|>{\centering\arraybackslash}p{3.93cm}|}
\hline
\textbf{Total parameters} & 95434 (372.79 KB) \\
\hline
\textbf{Trainable parameters} & 94986 (371.04 KB) \\
\hline
\textbf{Non-trainable parameters} & 448 (1.75 KB) \\
\hline
\end{tabular}

\end{table}

\begin{figure*}[!t]
\centering
\begin{minipage}[t]{.45\textwidth}
    \centering
    \includegraphics[width=\textwidth]{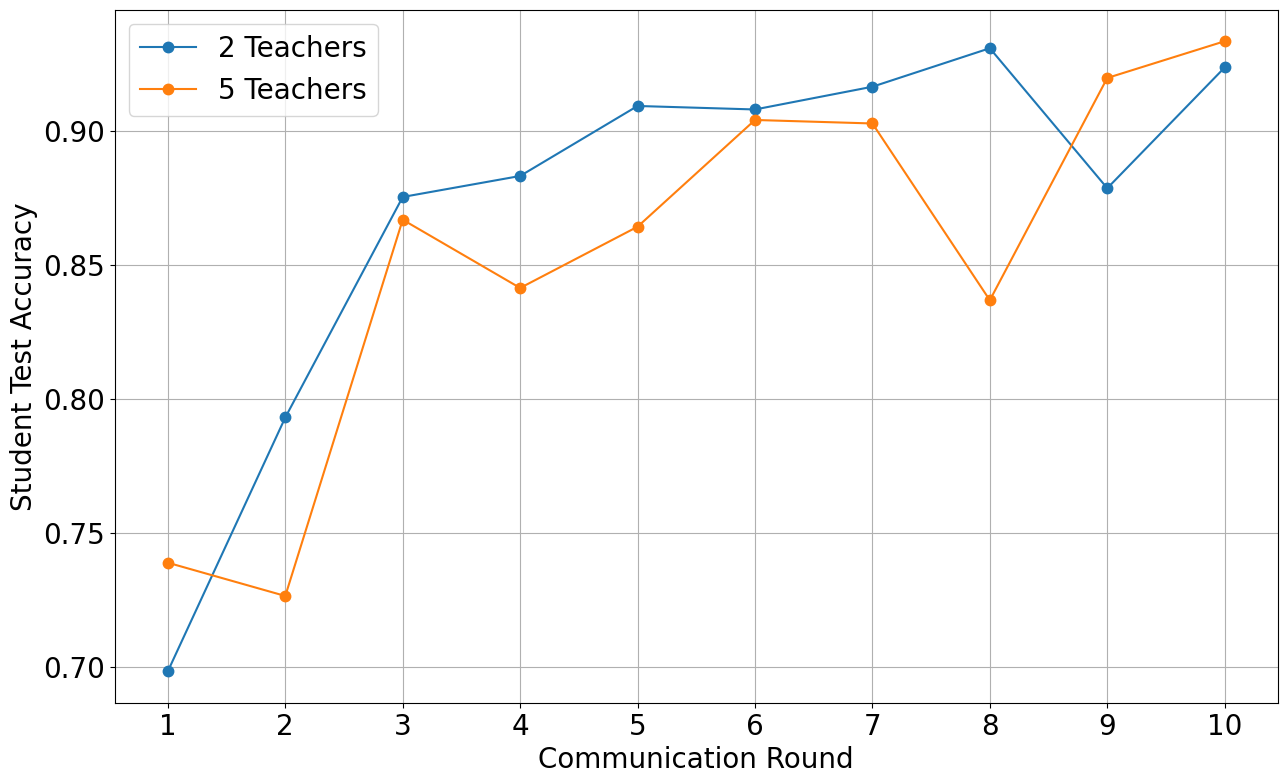}
    \caption{Student model accuracy after 10 communication rounds on non-IID data with 2 and 5 teachers.}
\end{minipage}
\hfill
\begin{minipage}[t]{.45\textwidth}
    \centering
    \includegraphics[width=\textwidth]{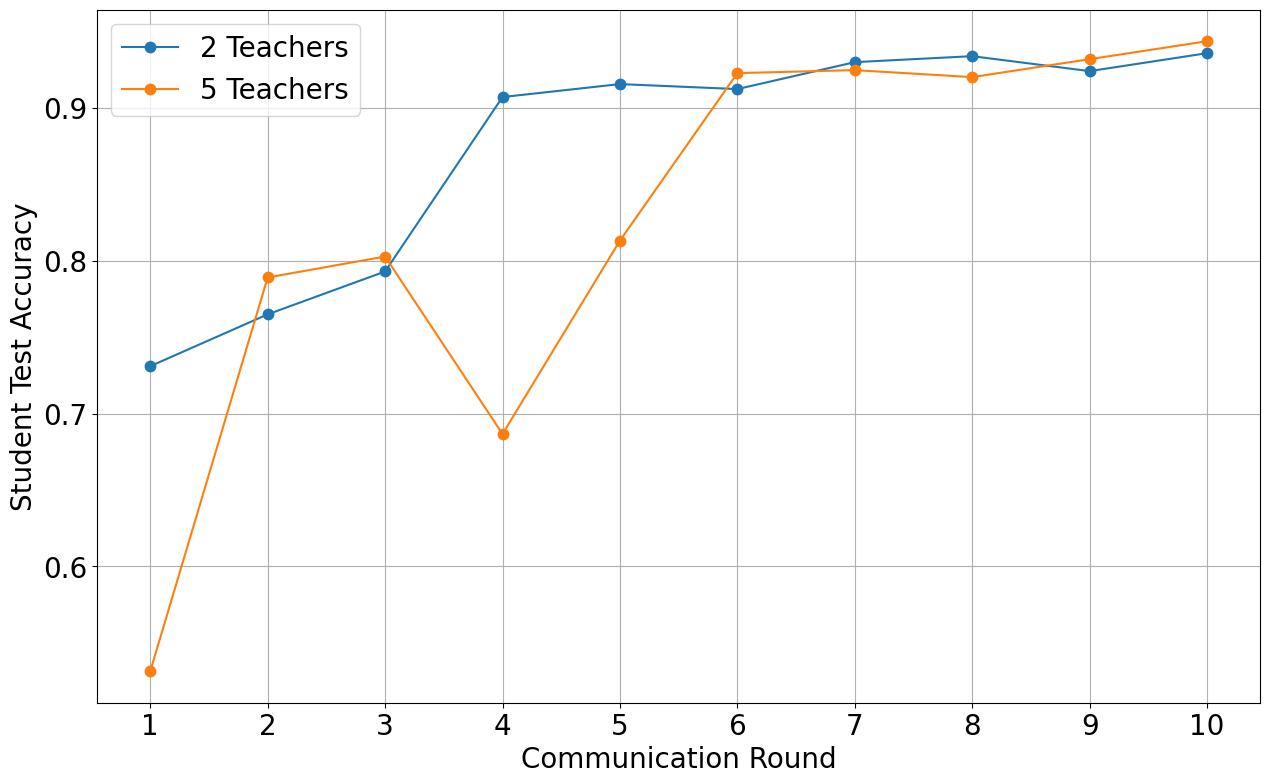}
    \caption{Student model accuracy after 10 communication rounds on IID data with 2 and 5 teachers.}
\end{minipage}
\end{figure*}

\begin{figure*}[!t]
\centering
\begin{minipage}[t]{.45\textwidth}
    \centering
    \includegraphics[width=\textwidth]{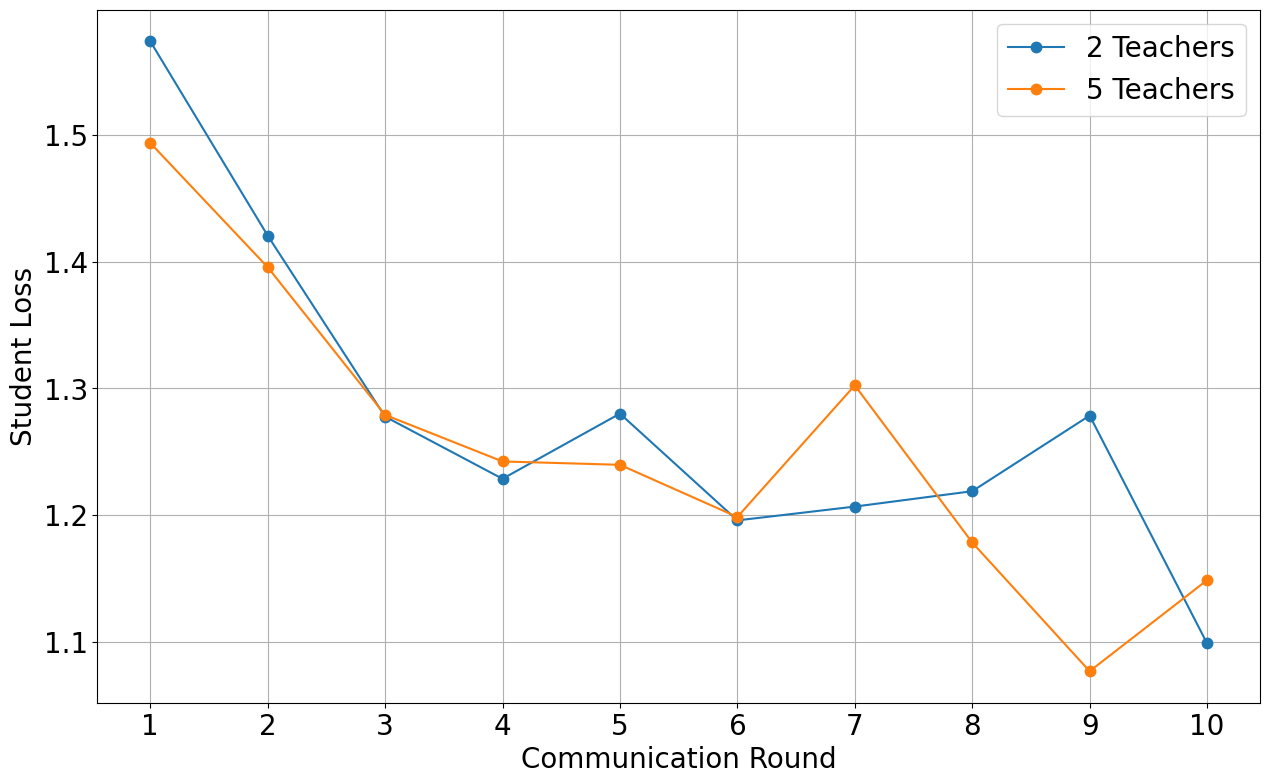}
    \caption{Student model loss after 10 communication rounds on non-IID data with 2 and 5 teachers.}
\end{minipage}
\hfill
\begin{minipage}[t]{.45\textwidth}
    \centering
    \includegraphics[width=\textwidth]{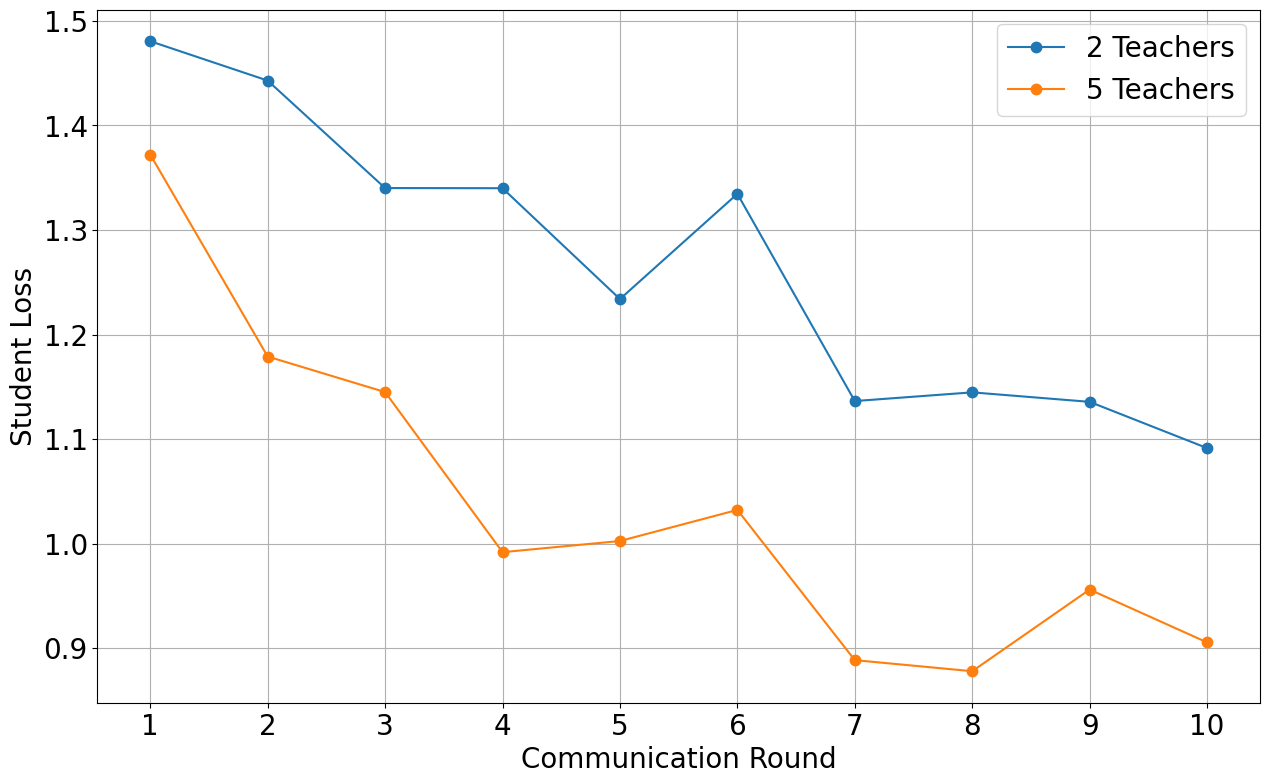}
    \caption{Student model loss after 10 communication rounds on IID data with 2 and 5 teachers.}
\end{minipage}
\end{figure*}
\begin{table*}[!t]
\centering
\caption{Performance comparison between FedBrain-Distill and an FL setting with 2 teachers/clients.}
\begin{tabular}{|c|c|c|c|c|c|c|c|c|c|}
\hline
\multirow{2}{*}{Method} & \multirow{2}{*}{Total Rounds} & \multicolumn{2}{c|}{Round 10 Accuracy (\%)} & \multicolumn{2}{c|}{Last Round Accuracy (\%)} & \multicolumn{2}{c|}{\makecell{Upload Communication \\ Cost per Round (Mb)}} & \multicolumn{2}{c|}{\makecell{Download Communication \\ Cost per Round (Mb)}} \\ \cline{3-10} 
                        &                                        & IID                & non-IID            & IID                 & non-IID             & IID                                   & non-IID                                    & IID                & non-IID            \\ \hline
FedBrain-Distill        & 10                                & 93.60            & 92.36            & 93.60            & 92.36             & 0.11 & 0.11                                    & 0.36           & 0.36           \\ \hline
Viet et al. \cite{c14}  & 100                                & 94.12 & 49.39 & 98.53            & 90.69            & 161.28 & 161.28 & 80.64           & 80.64 \\ \hline
\end{tabular}
\label{tab:mytable}
\end{table*}

\begin{table*}[!	t]
\centering
\caption{Performance comparison between FedBrain-Distill and an FL setting with 5 teachers/clients.}
\begin{tabular}{|c|c|c|c|c|c|c|c|c|c|}
\hline
\multirow{2}{*}{Method} & \multirow{2}{*}{Total Rounds} & \multicolumn{2}{c|}{Round 10 Accuracy (\%)} & \multicolumn{2}{c|}{Last Round Accuracy (\%)} & \multicolumn{2}{c|}{\makecell{Upload Communication \\ Cost per Round (Mb)}} & \multicolumn{2}{c|}{\makecell{Download Communication \\ Cost per Round (Mb)}} \\ \cline{3-10} 
                        &                                        & IID                & non-IID            & IID                 & non-IID             & IID                                   & non-IID                                    & IID                & non-IID            \\ \hline
FedBrain-Distill        & 10                                & 94.38 & 93.34            & 94.38 & 93.34             & 0.29 & 0.29                                    & 0.36           & 0.36           \\ \hline
Viet et al. \cite{c14}  & 100                                & 95.45 & 83.49 & 98.20            & 95.59            & 403.21 & 403.21                                   & 80.64           & 80.64 \\ \hline
\end{tabular}
\label{tab:mytable}
\end{table*}
\raggedbottom
\section{Experimental Results}
All our experiments were implemented in the Google Colab Pro environment using Python 3 with TensorFlow and Keras libraries. We used Figshare brain tumor dataset to train the student model and the teacher models. This dataset contains 3064 tumor images with three different classes of tumors;  meningioma, glioma and pituitary tumor \cite{c20}. We used Dirichlet distribution by setting the $\alpha$ parameter to 10000 and 0.5 to achieve the desired IID and non-IID data, respectively. Afterwards, we partitioned the private datasets for the teachers so that 80\% of the original data is used for the training set. This ratio holds true for both IID and non-IID data. As for the public dataset, 50\% of the original dataset was used to create a public train set. And finally for the test set, we used 50\% of the original data to create a separate test set in order for the student model to be evaluated on. We used 2 and 5 teachers in two different settings separately for 10 communication rounds. Also, we compared FedBrain-Distill with the work of Viet et al. \cite{c14}, who utilized VGGNet16 in an FL setting for tumor classification on Figshare brain tumor dataset. This makes it a fair comparison with a benchmark method since the architecture of the pretrained model as well as the training dataset are the same. All the teachers as well as the student model were trained on 5 local epochs for FedBrain-Distill. All the implementation details are accessible through our Github repository\footnote{\url{https://github.com/russelljeffrey/FedBrain-Distill}}. As it can be seen from figure 4 and figure 5, the student accuracy on the test set increases over 10 communication rounds on both IID and non-IID data with 2 and 5 teachers. However, with IID data, especially when the number of teachers is 2, the convergence seems to be smoother than when teachers are trained on non-IID data. Figure 6 and figure 7 show the gradual decrease of the student total loss over 10 communication rounds. Just like the accuracy, the student total loss is smoother on IID data. Table 2 demonstrates the comparison between FedBrain-Distill that utilizes 2 teachers and FL setting with 2 clients. FedBrain-Distill outperforms Viet et al. approach when data is non-IID. Even when their model is trained after 100 rounds of training, it still lags behind FedBrain-Distill. The communication of FedBrain-Distill is remarkably far lower for both the upload and download in comparison with Viet et al. approach.
Table 3 shows the comparison between FedBrain-Distill that utilizes 5 teachers and FL setting with 5 clients. As it can be seen, FedBrain-Distill performs better for the first 10 rounds when it comes to non-IID data. However, FedBrain-Distill fails by almost 2\% compared to the Viet et al. approach that used 100 rounds for training the global model. Compared to FedBrain-Distill, their approach is substantially expensive in terms of communication cost and convergence time for non-IID data. Work of Viet et al. is already consuming 403 Mb just to upload model parameters in one round. This becomes a challenge when number of clients increases in an FL setting. Also, their approach achieves only 2\% more accuracy on non-IID data after 90 more communication rounds. Given the potential that our approach shows, FedBrain-Distill can be further optimized in order to achieve better accuracy on both IID and non-IID data. 
Overall, we believe FedBrain-Distill improvement can be implemented by taking data augmentation and other statistical approaches into account, specifically when the teachers’ soft labels are aggregated. Other approaches that can be investigated to improve the student model accuracy are utilizing different architectures for the student model as well as changing the temperature $T$ in the softmax function or changing the $\alpha$ parameter to alter the loss ratio between the distillation and the student loss. Additionally, using multiple teachers with different pretrained models like U-Net or ResNet can be investigated to see the improvement of FedBrain-Distill.
\section{Conclusion}
Tumor classification has always been at the center of the attention in medical institutions. One of the most efficient approaches is taking advantage of DL approaches to classify tumor types in MRI images. On the other hand, DL approaches violate users' privacy by sharing the data across different medical institutions. We introduced FedBrain-Distill in order to carry out brain tumor classification more optimally. FedBrain-Distill not only maintains the users’ privacy, but also addresses some of the shortcomings of FL settings. These shortcomings include communication efficiency, dependency on model architecture and convergence time. FedBrain-Distill showed promising results against non-IID teachers. In addition, FedBrain-Distill can still be optimized by implementing different strategies, some of which include taking different student model architectures under consideration, changing the temperature $T$ in the softmax function, changing $\alpha$ parameter in the total loss formulation and finally using various other pretrained models like U-Net or ResNet for training the teachers.



%

\end{document}